\pdfoutput=1

\documentclass[11pt]{article}

\usepackage{naacl2022}

\usepackage{times}
\usepackage{latexsym}

\usepackage{todonotes}
\usepackage{tikz}
\usetikzlibrary{arrows,shapes,fit}
\usepackage{booktabs}

\usepackage[T1]{fontenc}

\usepackage[utf8]{inputenc}

\usepackage{microtype}

\title{Augmenting Poetry Composition with Verse by Verse}

\author{
 David Uthus \and Maria Voitovich \and R.J. Mical \\
 Google Research \\
  {\texttt{\{duthus, mvoitovich, gameman\}@google.com}} \\
}

\begin{document}
\maketitle
\begin{abstract}
We describe Verse by Verse, our experiment in augmenting the creative process of writing poetry with an AI.
We have created a group of AI poets, styled after various American classic poets, that are able to offer as suggestions generated lines of verse while a user is composing a poem.
In this paper, we describe the underlying system to offer these suggestions.
This includes a generative model, which is tasked with generating a large corpus of lines of verse offline and which are then stored in an index, and a dual-encoder model that is tasked with recommending the next possible set of verses from our index given the previous line of verse.
\end{abstract}


\section{Introduction}

There has been a lot of growing interest in poetry generation \cite{oliveira2017}.
Some of these approaches have even shown quality approaching that of humans \cite{lau2018}.
However, much of this has been in the view of letting an AI write a full poem by itself, thus writing in a closed system.
Only recently have some approaches started to explore human interaction when composing a poem \cite{ghazvininejad2016, ghazvininejad2017, oliveira17b, zhipeng2019}.

Verse by Verse\footnote{\url{https://sites.research.google/versebyverse/}} is our experiment in augmenting the creative process of poetry composition with an AI.
Unlike past approaches that focused on generating a full poem, we are interested on how we can use AI to offer suggestions to a user as they compose a poem.
This is a much more challenging task, as one needs be able to offer suggestions with minimal latency while meeting constraints of the poem structure and handle the challenges of user input.
Additionally, to make this a more educational experience, we wanted to generate the verses in the style of various classic American poets.

In this paper, we describe the underlying system that powers Verse by Verse.
Our main contributions are:
\begin{itemize}
    \item A novel approach using multiple models that allows us to split local verse knowledge (how to generate a line of verse) and global poem knowledge (what line of verse would best follow a previous line of verse).
    \item A novel way of determining rhyme phonemes for verses that is robust with user input.
    \item The first approach that we know that incorporates techniques to help reduce possible learned biases within a poetry system.
\end{itemize}

\section{Verse by Verse Overview}

\begin{figure*}
  \centering
  \includegraphics[width=.8\textwidth]{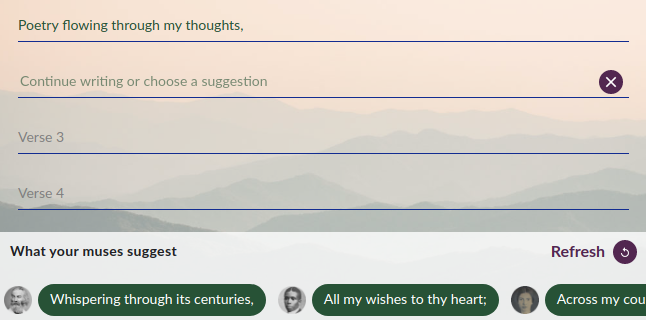} 
  \caption{UI of Verse by Verse, with a user composing a poem and the AI making suggestions.}
  \label{fig:pic}
\end{figure*}


As mentioned, Verse by Verse is an interactive application that allows users to compose a poem while getting suggestions from the system.
To use this application, users first pick a few classic American poets to act as their muses.
They will then pick the structure of the poem (quatrains, couplets, or free verse), and optionally syllable count and rhyme schema (when applicable).
Afterwards, they can begin to compose a poem.

While the user composes a poem, the poets will make suggestions of next possible lines of verse given the previous verse (as shown in Figure \ref{fig:pic}).
Users may either use these suggestions (including being able to edit the suggestions to make them more personal) or continue writing verses of their own.
This goes on until a user is satisfied with their poem, in which they can then optionally add a title\footnote{We had initially designed the system to start with a poem title, but feedback from our initial user subject studies showed that our poet enthusiasts preferred adding a title after a poem had been written. Having the title first made users feel forced to fit the poem to the title, while having the title last allowed them more freedom of creativity during composition.} and save the final poem as text or as an image.

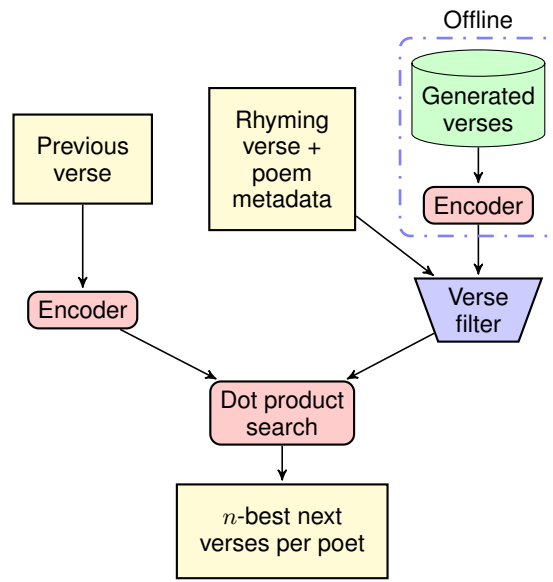
\begin{figure}
\centering
\begin{tikzpicture}[
  font=\sffamily\footnotesize,
  every matrix/.style={ampersand replacement=\&,column sep=0.3cm,row sep=0.5cm}, 
  source/.style={draw,thick,fill=yellow!20,inner sep=.3cm},
  process/.style={draw,thick,rounded corners,fill=red!20},
  filter/.style={trapezium,draw,thick,trapezium angle=110,fill=blue!20},
  sink/.style={draw,cylinder,shape border rotate=90,aspect=0.2,fill=green!20},
  dots/.style={gray,scale=2},
  to/.style={->,>=stealth',shorten >=1pt,semithick,font=\sffamily\footnotesize},
  every node/.style={align=center}]

  
  \matrix{
    \&  \& \node[sink] (data) {Generated\\verses}; \\[-13mm]
    \node[source] (input) {Previous\\verse}; \& \node[source] (rhyme_verse) {Rhyming\\ verse + \\ poem\\ metadata};  \&  \\[-11mm]
     \&  \& \node[process] (encode2) {Encoder}; \\[2mm]
    \node[process] (encode) {Encoder}; \& \& \node[filter] (rhyme) {Verse\\filter}; \\
    \& \node[process] (search) {Dot product\\search}; \& \\
    \& \node[source] (output) {$n$-best next \\ verses per poet}; \& \\
  }; 

  \draw[to] (rhyme_verse) -- (rhyme);
  \draw[to] (input) -- (encode);
  \draw[to] (encode) -- (search);
  \draw[to] (data) -- (encode2);
  \draw[to] (encode2) -- (rhyme);
  \draw[to] (rhyme) -- (search);
  \draw[to] (search) -- (output);
  
  \tikzset{blue dotted/.style={draw=blue!50!white, line width=1pt,
                               dash pattern=on 1pt off 4pt on 6pt off 4pt,
                               inner sep=1.5mm, rectangle, rounded corners}};

  \node (dotted box) [blue dotted, fit = (data) (encode2)] {};
  \node at (dotted box.north) [above, inner sep=1mm] {Offline};
  
\end{tikzpicture}

\caption{Overview of underlying system that handles user input and suggests next possible lines of verse.}
\label{fig:overview}
\end{figure}

Figure \ref{fig:overview} shows an overview of how we suggest verses to the user.
Our system first receives from the user as input: the previous verse, poem structure metadata (such as syllable count and selected poets), and, if needed, a verse to rhyme with.
When a rhyming verse is provided, the system will find the rhyming syllables for this verse.
The rhyming syllables along with the poem structure metadata will then be used as filters on the generated verse suggestions.
With the previous verse input, the system will then encode the verse using a feed forward network.
This encoding will be used in a search against pre-generated and pre-encoded verses, taking the dot product of each pair of encodings.
It will then output a list of the $n$-best\footnote{The value of $n$ is controlled by the UI, which considers two factors: whether the user is on desktop or mobile (we can show more suggestions when viewing on desktop) and how many poets the user has selected to act as their muses.} possible verses per poet to suggest as the next verse based on the dot product scoring.

The next few sections will cover the various parts of the system: verse generation, verse retrieval, and determining rhyme syllables.

\section{Offline Verse Generation}\label{verse_generation}

We generate our verses offline and store them for later retrieval, which differs from past approaches of poetry generation.
This allows for faster serving \cite{henderson2017}, especially when used in a dual encoder network as described in Section \ref{prediction}.

\begin{figure*}
\centering
\begin{tikzpicture}[
  font=\sffamily\footnotesize,
  every matrix/.style={ampersand replacement=\&,column sep=0.3cm,row sep=0.4cm},   
  source/.style={draw,thick,fill=yellow!20,inner sep=.3cm},
  model/.style={draw,thick,rounded corners,fill=red!20},
  process/.style={trapezium,draw,thick,trapezium angle=110,fill=yellow!20},
  sink/.style={draw,cylinder,shape border rotate=90,aspect=0.2,fill=green!20},
  dots/.style={gray,scale=2},
  to/.style={->,>=stealth',shorten >=1pt,semithick,font=\sffamily\footnotesize},
  every node/.style={align=center}]

  \matrix{
     \& \node[sink] (ws) {Whitman\\corpora}; \& \node[model] (w1) {Whitman\\model}; \& \node[process] (w2) {Generate\\verses}; \& \node[process] (w3) {Filter\\verses}; \& \node[process] (w4) {Add\\metadata}; \&  \\
     \node[sink] (data) {Poetry\\corpora}; \& \node[model] (t1) {Transformer\\model}; \&  \&  \& \& \& \node[sink] (data2) {Generated\\Verses};  \\
     \& \node[sink] (ds) {Dickinson\\corpora}; \& \node[model] (d1) {Dickinson\\model};  \& \node[process] (d2) {Generate\\verses};  \& \node[process] (d3) {Filter\\verses}; \& \node[process] (d4) {Add\\metadata}; \&  \\
  };

  \draw[to] (data) -- (t1);
  \draw[to] (t1) -- (w1);
  \draw[to] (ws) -- (w1);
  \draw[to] (t1) -- (d1);
  \draw[to] (ds) -- (d1);
  \draw[to] (w1) -- (w2);
  \draw[to] (w2) -- (w3);
  \draw[to] (w3) -- (w4);
  \draw[to] (w4) -- (data2);
  \draw[to] (d1) -- (d2);
  \draw[to] (d2) -- (d3);
  \draw[to] (d3) -- (d4);
  \draw[to] (d4) -- (data2);

\end{tikzpicture}

\caption{Overview of how we generate our lines of verses offline. We begin with the full corpora of English poetry and train a transformer model. We then copy this model and fine tune it for each of our poets on their individual corpora, using Whitman and Dickinson as examples here. These models are then used to generate novel verses, which are filtered for quality and amended with metadata. All these generated verses are then added to our generated verses index, which is used for serving lines of verse to our uses.}
\label{fig:generation}
\end{figure*}
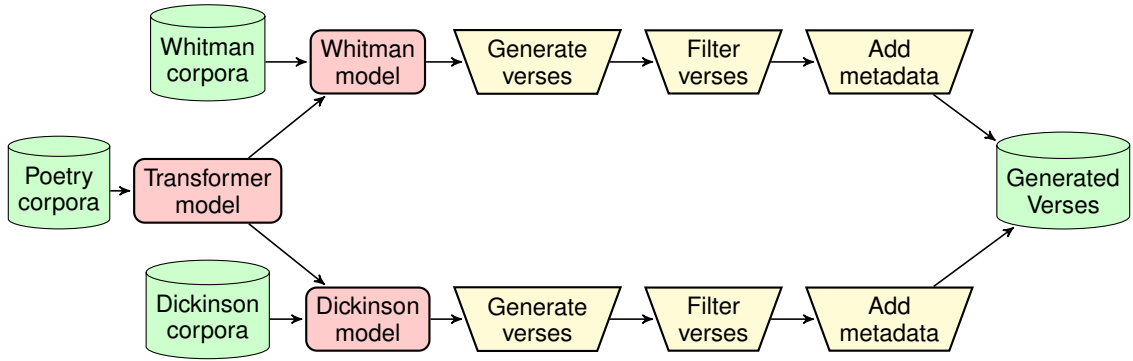

Our verse generation is done in a pipeline composed of multiple steps.
Figure \ref{fig:generation} shows an overview of this.
It takes original poetic sources and creates new verses (Section \ref{generate}); then filters out poorly-generated verses (Section \ref{quality}); and finally adds metadata for each verse such as the rhyme syllables and syllable count (Section \ref{metadata}).

\subsection{Generating Novel Verses}\label{generate}

In our approach, we present users the option to choose from 22 American poets to act as their muses.
These poets are restricted to those in which there is substantial enough material available to use that is no longer under copyright, with most material found on Project Gutenberg\footnote{http://www.gutenberg.org/}.

\subsubsection{Architecture and Training}

We use a decoder-only Transformer model \cite{vaswani2017} for generating these verses.
This model is trained to predict the next single subword token given the previous tokens in a line of verse.
It is composed of 8 multi-head attention layers with 8 heads each.
The layers had a hidden dimensionality of 128 and feed-forward dimensionality of 512.

We first pretrain the model on a large corpus (1,116,297 lines of verse) of English-speaking poems from Project Gutenberg, including the above mentioned 22 American poets.
This is done for 400 epochs.
Following this, we make 22 copies of the model and fine-tune each one on a given poet, training for 50 epochs per poet.
This fine-tuning then allows us to capture the style of each poet.
For both phases of training, we use a batch size of 128, dropout rate of 0.1, and the same learning rate as described in the Transformer paper \cite{vaswani2017}.

\subsubsection{Generation}

After the generative models have been trained, we next start generating all feasible lines of verse per poet.
This involves taking a set of starting tokens and then extending with all the suggestions of the model given a certain threshold is met.
The set of starting tokens is composed of the original starting tokens of the lines of verse by a given poet combined along with tokens that were common across the 22 poets.
The extra starting tokens are particularly beneficial for poets whose corpus is small and might not have as many tokens to start a verse with.
But to help to avoid introducing uncommon tokens that may be part of a poet's style, we restrict to tokens that have been used by at least 12 poets.

For each partial verse (any verse that does not yet contain an end-of-line token), we expand it by considering all tokens whose normalized score (probability of being the next token normalized against the maximum probability) are above a threshold of 0.925. 
An expansion that results in an end-of-line token will then be included in that poets' generated corpus.
Any incomplete verses will be considered for another iteration of expansion.
This continues on for 10 iterations.
To help contain the exponential growth as we generate the lines, for each iteration, we only carry over the 100M best partial verses seen that iteration.
This is determined by summing the scores seen so far for a given partial verse.


\subsubsection{Quality vs Quantity}

As mentioned in the previous section, we had used a threshold of 0.925 for the verse generation.
We had experimented with different values, and found this to give us the best balance of quantity vs quality.
Intuitively, having a higher threshold would result in much better quality of verses, though allowing for only a smaller set of generated verses.
And for a closed system, where topics can be more restricted, this would have sufficed.
But as we need to handle any possible topic presented by the user, we needed to loosen quality in order to allow for a wider variety of verses.

\subsection{Quality Control Filtering}\label{quality}

After we have generated our collection of verses, we then run them through various filters to remove those of poor quality (especially as discussed in previous section we have lowered the quality threshold to allow for a wider variety of verses).
This includes: making sure parenthesis and quotation marks are balanced, filtering out verses of syllable counts not supported in the application, removing verses which contain words that we do not want to serve to the user (e.g., offensive words), and filtering out any verse that matches one of the original verses written by the given poet.

An additional filter we implemented is filtering by part-of-speech.
Using the large corpus of English-speaking poems, we go through each line of verse and get a POS ``fingerprint,'' which is a concatenation of POS tags representing a line of verse.
Then, for every generated verse, we check to see if that line of verse's POS matches that of one of the fingerprints from the original verses.
If so, we keep the line of verse, otherwise it is removed from our collection.
The reasoning behind this is that since we are doing a deep search of many possible verses with our generative model, it will sometimes generate lines of verse of very poor grammar.\footnote{While poets may compose lines of verse that purposely break the rules of grammar, we are more focused on filtering out lines of verse that are unreadable due to their poor grammar.}
By utilizing the POS used by our real poets, we can then help to improve the quality of the generated verses.\footnote{Alternatively, we had experimented with language model classifiers prior to implementing the POS fingerprint filtering. These classifiers did not work very well, oftentimes removing too many good verses or allowing too many poor quality verses to pass through, especially for our poets with small bodies of work available.} 






After the filtering, we are left with a total of 26.9M generated verses for our 22 poets.
But as our poets all have different styles, along with a different amount of available past works available, some poets will have a resultant larger set of generated verses than others, ranging from 60K for our smallest to  8.3M for our largest.

\subsection{Metadata}\label{metadata}

All generated verses that are of good quality are finally labeled with metadata.
This metadata includes the poet source this was generated from, syllable count and rhyming phoneme (to be discussed later in Section \ref{rhymes}), and any other fields we may need to filter upon for serving to our users.

\section{Next Verse Prediction}\label{prediction}

We use a dual encoder network architecture for suggesting the next line of verse of a poem.
We will discuss training of the network, the indexing of possible verses, and the retrieval of verses.

\subsection{Dual Encoder Model}

We use a dual-encoder architecture that is similar to what was used in Gmail's Smart Reply \cite{henderson2017}.
In the original work, the authors would encode the user input with one encoder and all possible replies with the other encoder.
In our model, one encoder is used to encode a parent (previous) verse and the other encoder is used to encode a child (next) verse.
Then, same as in the original work, the model optimizes for a given verse’s dot-product score with the true following verse to be higher than with random negatives from the batch.

Our network does differ though from the original work with respect to the composition of the encoders.
For the two encoders in our network (as shown in Figure \ref{fig:overview}), they both take in an input and feed that into a SentencePiece model \cite{kudo2018}, consisting of a vocab size of 128K.
This then feeds into a set of Transformer layers \cite{vaswani2017}.
The Transformer consists of 4 layers, each with 4 attention heads, a hidden size of 1024, and a feed-forward size of 4096.
Finally, these then feed into a set of 2 fully-connected layers, with ReLU activation on the first layer and Softsign activation on the second.
These deep layers consist of a hidden size of 500 each.
In terms of weights, the Transformer layers for the two stacks share weights while the fully-connected layers do not.







\subsection{Training}

We use two collections of data for training data.
One is a mixture of poems (such as those used for poetry generation) and other similar mediums, which we call \textit{poetic}.
This set's purpose is to train the model to predict the next line of verse given the previous.
The other is composed of comments from internet discussion forums, which we call \textit{comments}.
For this, we train to predict a comment given the previous comment.
Doing so allows us to expose the model to a larger vocabulary and more noisy data than what would normally be seen in the \textit{poetic} corpus, which is important when dealing with user input.

To train the dual encoder, we first pretrain the model on our \textit{comment} data for 20M steps with a learning rate of 0.01.
After this, we will finetune the model on the \textit{poetic} corpus for an additional 10M steps with a learning rate of 0.001.
We use dropout for both the Transformer attention and ReLU layers of 0.1.
We use a training batch size of 100.
Additionally during training, we use the parent (previous verse or comment) as extra negative examples, which helped train the model not to repeat itself.

\subsection{Verse Indexing and Retrieval}

After we have trained the dual-encoder model, we can then use it to start to encode all our generated verses from the previous section.
Each generated verse will be encoded using the encoder for the child verse.
These are then stored in an index.
During retrieval, instead of using an exhaustive search across all possible verses, we use a hierarchical quantization approach for allowing for fast search \cite{guo2016,wu2017}.



When composing a poem, the system will receive the previous verse and various metadata for filtering, as shown in Figure \ref{fig:overview}.
We first encode the previous verse using the parent encoder.
We then take the dot product of this verse and all possible verses, filtering out verses based on what the user needs.
Afterwards, the system will return the $n$-best possible next verses.

In the end, this architecture allows us to do a lot of the expensive process offline and allows for fast retrieval and filtering when users are composing a poem.
Additionally, this adds the capability of filtering verses by their respective metadata, so that we can match the requirements of what a user desires for the structure of their poem.

\section{Rhymes and User Input}\label{rhymes}

As we allow users to enter their own verses or edit candidate verses, we then have to take this into account for next-verse suggestion when dealing with rhyme.
In many past approaches that are generating a poem in full, they can use various heuristics to help meet requirements for rhyme, such as restricting what words are available.
Or in some cases, such as with Deep-speare, learn a model for rhyme \cite{lau2018}.
Since we were creating an interactive approach, we then had to take a different route for dealing with rhymes.

\subsection{Text Normalization}\label{phonemes}

For rhyme syllables (and syllable count), we initially used the CMU pronunciation dictionary\footnote{http://www.speech.cs.cmu.edu/cgi-bin/cmudict}, which has been used in past approaches such as \citet{ghazvininejad2016} and \citet{hopkins2017}.
This was unfortunately problematic -- its use is limited when dealing with words with multiple pronunciations (e.g., past and present tense of ``read''), and failed when handling irregular spelling and out-of-dictionary words both from what poets used in their writings (e.g., ``W'en daih's chillun in de house,'' by Paul Laurance Dunbar) and when handling user input.
We also considered training a model for rhyme, similar to what was done for Deep-speare \cite{lau2018}.
While this would help alleviate some of the dictionary issues, it would still be fragile when handling user input. 

To overcome these issues, we used the Kestrel text normalization system \cite{ebden2015} for determining the rhyming syllables and syllable counts of a verse.
It is able to determine correct pronunciations of words like ``read'' with respect to tense, and is able to suggest phonemes for out-of-dictionary words.
Furthermore, it can handle more extreme situations, such as ``In my pocket there is \$.50''.
In this case, the system is able to understand that it needs to find a word that rhymes with ``cents''.

\subsection{Perfect and Imperfect Rhymes}

This work uses both perfect and imperfect rhyming.
For the imperfect rhyming, we loosely follow the steps as described by \citet{ghazvininejad2016}, with slight modifications to accommodate the difference between their use of the CMU dictionary and our use of Kestrel.

Expanding beyond their work, we also allow for imperfect rhymes on single-syllable words.
For this, we find similar consonant phonemes for the last phoneme of the word where the logs-odd scoring is 0 or greater from the work by \citet{hirjee2010}.

When a user is composing a poem and we need to suggest a rhyming line of verse, the system will attempt to show a mixture of both perfect and imperfect rhyming verses.
Only if it is unable to find any verses that rhyme, a possibility given the wide range of possible inputs, it will then show non-rhyming verses.

\section{Evaluation}

\begin{table}
    \centering
    \begin{tabular}{ c | c | c} \toprule
       & Human & Verse by Verse \\ \midrule
       Judged human & $82.7\%$ & $47.0\%$ \\
       Readability & 3.8 & 2.9 \\
       Relevance & 3.9 & 3.2 \\
       Evocative & 3.4 & 2.7 \\
       Aesthetic & 3.7 & 3.0 \\ \bottomrule
    \end{tabular}
    \caption{Human evaluations comparing poems written by classic poets with those generated with Verse by Verse. ``Judged human'' represents the percentage of quatrains that the evaluators had judged as having been written by a human. The four proceeding metrics were judged on a scale of 1-5.}
    \label{tab:human}
\end{table}

We ran comparative evaluations of Verse by Verse against poems written by classic poets.
While Verse by Verse is meant to be used in an interactive setting to aid a user in writing a poem, we felt it was still worth evaluating how well it works on its own in writing a poem given a first line of verse.

To do so, we gathered a collection of 100 quatrains written by the 22 classic poets.
Then, for each quatrain, we would take the first line of verse, and use that as the first line of verse for Verse by Verse.
It would then take the top suggestion (using the same poet as that who wrote the poem), to pick the subsequent 3 lines.
When possible, it would try to follow an ABAB rhyme pattern.

We built upon the work of \citet{hopkins2017} for evaluating.
We would show evaluators one poem at a time.
They then needed to classify if the poem is human- or AI-written (they are shown the full quatrain and asked to evaluate on the last 3 lines), and rate on a scale of 1-5 for readability (to what extent is the quatrain easy to read? does it make sense?, relevance (given the first line, how relevant are the subsequent lines of verse?), evocation (how much does the quatrain evoke emotion when reading it?), and aesthetic (how much does the quatrain sound nice to read, such as in rhythm?).
Each poem was evaluated by 3 evaluators.

Table \ref{tab:human} shows the results of these evaluations.
As shown, while Verse by Verse does not do as well as the poems written by classic poets, it still was able to do well enough.
More importantly, almost half the poems written by Verse by Verse were thought of to have been written by humans, which shows the feasibility of our approach.
A couple of the highly rated poems can be seen in Figure \ref{fig:examples}.

\begin{figure}
    \centering
    \textit{Her eyes, twin pools of mystic light,}
    
    Forever in her radiance white—,
    
    She sought the bosom of the Night.
    
    Away it came, that mystic sight!
    
    \vspace{5mm}
    
    \textit{Whether I travel by land or by sea,}
    
    Just while I travel with its fairy tide,
    
    Leaving a gleam that I may never see,
    
    Although I travel close upon your side.
    \caption{Example quatrains rated highly by evaluators. The first line is by a poet and the subsequent 3 lines are generated by Verse by Verse.}
    \label{fig:examples}
\end{figure}


\section{Related Work}

Poetry generation is a growing field of research, with many diverse approaches for generating full-length poems of various forms \cite{oliveira2017}.
Some related areas to touch upon are user interaction and verse generation.

\subsection{Interactive Generation}

There have been some recent work that have looked at interactive approaches to composing poetry or song lyrics.

Jiuge \cite{zhipeng2019} is a similar interactive approach to writing Chinese poetry.
Users would input keywords, text or images, and from there the system would extract keywords to use within a generative model for writing the poems.
Users could then edit or make use of suggestions.

Hafez \cite{ghazvininejad2016,ghazvininejad2017} offered a variety of inputs for users to dictate how a poem was generated (e.g., topic; desired words; control for sentiment, alliteration, etc.), and then automatically generated a full poem given these inputs.
Users could then further tweak the controls until a poem was generated to their liking.
Underneath, given these set of input values, it would initiate with a candidate set of rhyming words, and then use a Finite State Acceptor to guide a Recurrent Neural Network for generating new verses.

Co-PoeTryMe \cite{oliveira17b}, which was built on PoeTryMe \cite{oliveira2012}, would generate full poems given some inputs (e.g., keywords, number of syllables).
Users were then allowed to edit lines and use suggested lines as seeds for further generation of suggestions.

DopeLearning  \cite{malmi2016} was focused on generating rap lyrics.
It allowed for interactive rap composition -- for each verse, a user could either pick from a list of candidates or enter their own input.
For determining its suggestions, DopeLearning treated their approach as an information retrieval task, ranking the best response given the previous verse.
DopeLearning was restricted though in only reused existing rap verses, as it did not generate any novel verses.

\subsection{Verse Generation}

There have been many different approaches to how a line of verse is generated.
Earlier works included template-based approaches \cite{colton2012,oliveira2012} while most recent works have been neural-based approaches \cite{ghazvininejad2016,ghazvininejad2017,hopkins2017,lau2018,vandecruys20,yi2018,zhang2014}.

As with recent approaches, ours is also considered a neural-based approach.
Our approach is closest to the work of \newcite{liao19}, which involved a Transformer-based approach using GPT.
Both their approach and ours used pre-training and fine-tuning of the models, though the type of data used differs.
Our model used poetry data for both phases, while their approach first pre-trained on a news corpus, then fine-tuned on Chinese poetry.
An additional difference is how the models are used -- they used their model for generating a whole poem while we use our model for offline verse generation of single lines of verse and instead rely on a dual-encoder model for determining the next line of verse in a poem.











\section{Conclusions}

We have described the underlying system of Verse by Verse.
It is composed of two primary models, one for verse generation and one for verse recommendation.
Results show that this approach works well for an interactive setting, generative novel verses that do well in human evals and meet the more challenging demands of human interaction.

\section*{Ethical Concerns}

As this system is intended to be deployed to a general audience of all ages, there are concerns of how the tool can accidentally suggest offensive verses.
We have taken some steps to help alleviate this: augmenting the training data and filtering out problematic verses.

\subsection*{Augmenting Training Data}

We have augmented some of the \textit{poetic} data to help reduce bias using the techniques described in \newcite{sheng20}.
In their work, they had used a style transfer model to augment some of the data to make the sentiment more positive, with particular focus for the case when the parent verse contained a demographic mention.
In do so, this then helps move the model to suggest verses of more positive sentiment when the previous verse of a poem contains a demographic mention.

As with their work, we augment all child verses that have parent verses containing demographic mentions and about $50\%$ of those without a parent verse containing a demographic mention.
While we followed much of their described approach, we use a different style transfer model though for our augmentations, using TextSETTR \cite{riley20} as a replacement.
TextSETTR was shown to yield better results in transforming sentiment while preserving fluency (important aspects for our work).
As described in the TextSETTR paper, we use the model that had been fine-tuned on English Common Crawl data.
To change sentiment, we gave the model 10 examplars each of negative and positive lines, and then used this to change the sentiment of negative lines of verse using the techniques described in \newcite{sheng20}.\footnote{
For positive examplars, we used:
``The food was great!'',
``I really loved it.'',
``Absolutely my favorite book.'',
``I am filled with love.'',
``The seas are calm.'',
``She delights me'',
``He understands me,'',
``My soul is full of light,'',
``The scene is full of heroes'',
``This cup of tea tastes delightful''.
For negative examples, we used:
``The food was awful!'',
``I really hated it.'',
``I regret reading this book.'',
``I am filled with hatred.'',
``The seas are violent'',
``She annoys me'',
``He ignores me,'',
``My soul is full of darkness,'',
``The scene is full of villans'',
``This cup of tea taste horrible''.
}

We note that even though we have changed some of the sentiment to make the system as a whole more positive, it does not prevent users from writing negative poetry.
If a user writes a negative verse, the system can still suggest negative verses.
Additionally, the system does allow users to edit suggestions, so a user can also edit a verse to make it more negative if that is their desire.

\subsection*{Verse Filtering}

We also filter out verses that can potentially be offensive.
This includes filtering out verses that contain obscene words (especially as what was acceptable in the past might not be acceptable today), along with verses that may contain groups of words that, when put together, can be offensive.

One of the advantages of our system, where we generate and store our verses offline in an index, is that it makes it easier to explore how the filters would impact what verses we have available.
We can see if certain grouping of words are present in the index, and if such, filter out such verses.
More importantly, this allows us to further check if filtering out a group of words may filter out too many verses that would not be offensive, and thus allow us to better refine the word filtering as needed.

\bibliography{naacl2022}




\end{document}